\documentclass[final]{cvpr}

\usepackage{times}
\usepackage{epsfig}
\usepackage{graphicx}
\usepackage{amsmath}
\usepackage{amssymb}
\usepackage{booktabs}
\usepackage{multirow}
\usepackage{nopageno}
\usepackage[font=small]{caption}
\usepackage[ruled,vlined]{algorithm2e}

\usepackage[pagebackref=true,breaklinks=true,colorlinks,bookmarks=false]{hyperref}

\def\cvprPaperID{1157} %
\def\confYear{CVPR 2021}

\newif\ifdraft
\drafttrue

\newcommand{\RR}[1]{\ifdraft {\color{cyan}{\bf #1}} \else {}\fi}

\newcommand{\MS}[1]{\ifdraft {\color{magenta}{\bf #1}} \else {}\fi}
\newcommand{\comment}[1]{}

\newcommand{\skdt}{S-KdT}
\newcommand{\fullskdt}{sparse Kendall-Tau}

\newcommand\mypara[1]{\vspace{1mm}\noindent\textbf{#1}}

\usepackage{xr}
\makeatletter
\newcommand*{\addFileDependency}[1]{
  \typeout{(#1)}
  \@addtofilelist{#1}
  \IfFileExists{#1}{}{\typeout{No file #1.}}
}
\makeatother

\begin{document}

\title{Landmark Regularization: Ranking Guided Super-Net Training in Neural Architecture Search}

\author{Kaicheng Yu\thanks{This work was partially done during an internship at Intel, and supported in part by the Swiss National Science Foundation.}\\
CVLab, EPFL\\
{\tt\small kaicheng.yu.yt@gmail.com}
\and
Ren\'{e}~Ranftl\\
Intelligent Systems Lab, Intel\\
{\tt\small rene.ranftl@intel.com}
\and
Mathieu Salzmann \\
CVLab, EPFL\\
{\tt\small mathieu.salzmann@epfl.ch}
}

\maketitle
\begin{abstract}
Weight sharing has become a de facto standard in neural architecture search because it enables the search to be done on commodity hardware. However, recent works have empirically shown a ranking disorder between the performance of stand-alone architectures and that of the corresponding shared-weight networks. This violates the main assumption of weight-sharing NAS algorithms, thus limiting their effectiveness. We tackle this issue by proposing a regularization term that aims to maximize the correlation between the performance rankings of the shared-weight network and that of the standalone architectures using a small set of landmark architectures. We incorporate our regularization term into three different NAS algorithms and show that it consistently improves performance across algorithms, search-spaces, and tasks.
\end{abstract}

\section{Introduction}

Modern algorithms for neural architecture search~(NAS) can now find architectures that outperform the human-designed ones for many computer vision tasks~\cite{liu2019autodeeplab,wu_fbnet:_2018,chen2019detnas,ryoo2020assemblenet}. A driving factor behind this progress was the introduction of parameter sharing~\cite{Pham2018}, which reduces the search time from thousands of GPU hours to just a few and has thus become the backbone of most state-of-the-art NAS frameworks~\cite{Bender2020tunas,yu2020bignas,Cai2020Once,luo2020seminas}. At the heart of all these methods is a shared network, a.k.a. super-net, that encompasses all architectures within the search space. 

\begin{figure}
    \centering
    \resizebox{\linewidth}{!}{\includegraphics{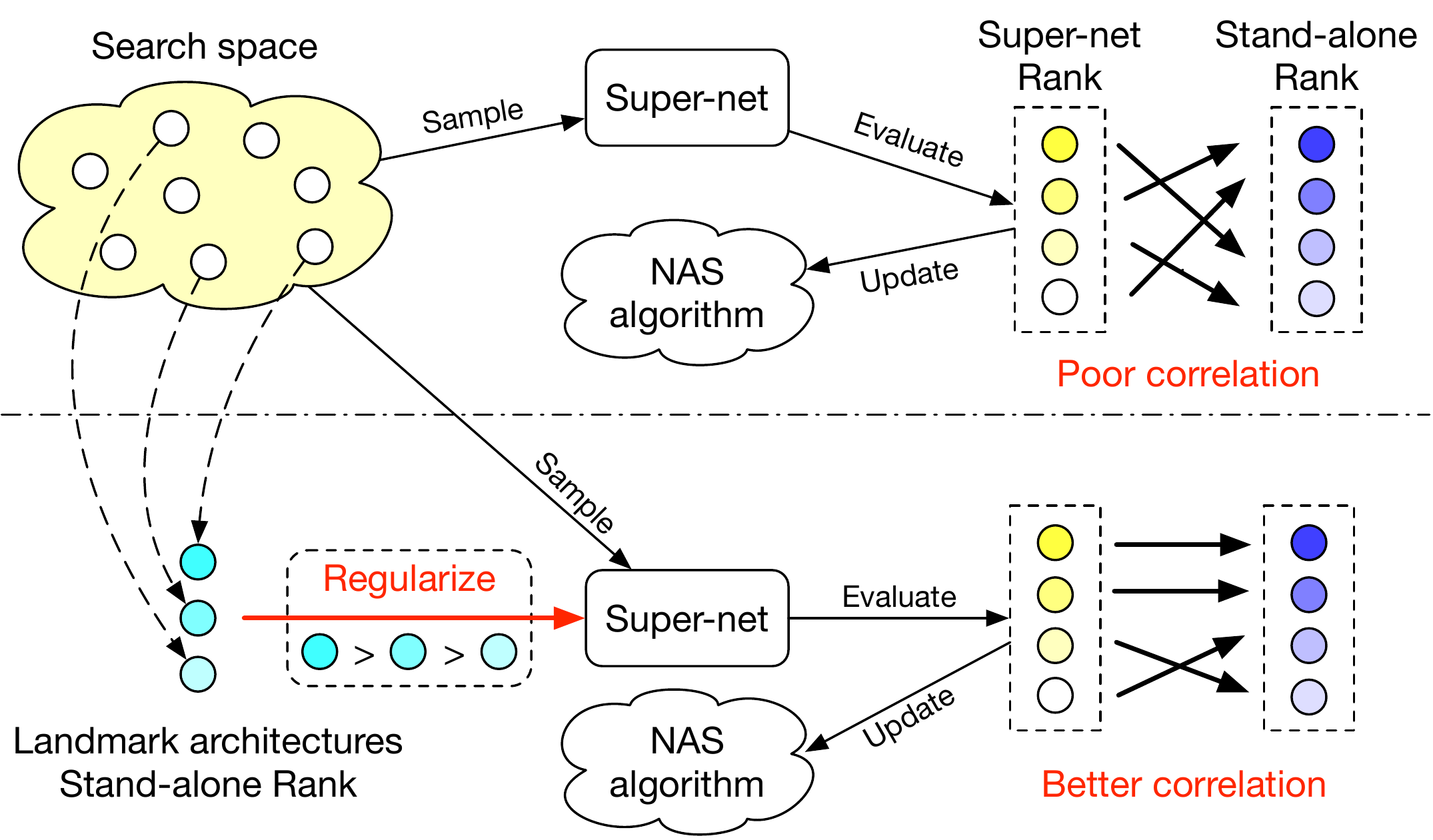}}
    \caption{
    Traditional super-net training leads to poor correlation between relative stand-alone performance and super-net performance (top). We sample landmark architectures and use their relative performance to guide training towards an improved ranking and show that this improves the search performance (bottom).
    }
    \label{fig:teaser}
    \centering
    \vspace{-0.2cm}
\end{figure}

To train the super-net, NAS algorithms essentially sample individual architectures from the super-net and train them for one or a few steps. The sampling can be done explicitly, with strategies such as reinforcement learning~\cite{Pham2018,cai2018proxyless}, evolutionary algorithms~\cite{guo_single_2019,wang2020neural}, or random sampling~\cite{yu2020evalnas,li2019random}, or implicitly, by relying on a differentiable parameterization of the architecture space~\cite{li2019improving,Liu2018darts,Cai2020Once,wu_fbnet:_2018,Zela2020Understanding}. Whether explicit or implicit, the underlying assumption of these methods is that the relative performance of the individual architectures in the super-net is highly correlated with the performance of the same architectures when they are trained in a stand-alone fashion. If this were the case, one could then safely choose the best individual architecture from the super-net after the search and use it for evaluation. However, this assumption was disproved in~\cite{yu2020evalnas,Zela2020NAS-Bench-1Shot1}, who showed a correlation close to zero between the two rankings on complex search spaces~\cite{ying2019bench,Liu2018darts}. The major reason behind this is fairly intuitive: To be optimal, different individual architectures should have different parameter values, which they cannot because the parameters are shared. 
Super-net training will thus \emph{not} produce the same results as stand-alone training. More importantly, there is no guarantee that even the relative ranking of the architectures will be maintained. 
While for simple, linear search spaces the ranking can be improved by using a carefully crafted sampling strategy~\cite{chu_fairnas:_2019,chu2020fairdarts},
addressing the ranking disorder for more realistic, complex search spaces remains an open problem~\cite{yu2020train}.

In this paper, we propose to explicitly encourage architectures represented by the super-net to have a similar ranking to their counterparts trained in a stand-alone fashion. As illustrated by Figure~\ref{fig:teaser}, we leverage a set of landmark architectures, that is, architectures with known stand-alone performance, to define a regularization term that 
guides super-net training towards this goal.
We show that a small set of landmark architectures suffices to significantly improve the global ranking correlation, so that the overall search procedure, including the independent training of the landmark architectures, remains tractable.

Our regularization term is general and does not make assumptions about the specific sampling algorithm used
for super-net training. As such, it can easily be combined with many popular weight-sharing NAS algorithms. We demonstrate this by integrating it into three different algorithms \cite{guo_single_2019,Luo2018,dong2019searching} that are representative of three different
categories of weight-sharing NAS algorithms: i) Algorithms that sample 
architectures from the super-net in an unbiased manner throughout the super-net training \cite{li2019random,yu2020evalnas,Bender2018,guo_single_2019,chu_fairnas:_2019}; ii) approaches that employ learning-based samplers, which are updated during the training based on the performance of the partially-trained super-net~\cite{Pham2018,li2019improving,Luo2018,wang2019alphax,zhao2020fewshot}; and iii) algorithms that rely on 
differentiable architecture search~\cite{Liu2018darts,cai2018proxyless,xie2018snas,nayman2019xnas,Xu2020PC-DARTS:}.

Our extensive experiments on CIFAR-10 and ImageNet show that landmark regularization significantly reduces the ranking disorder that occurs in these algorithms and that they are consequently able to consistently find 
better-performing architectures. To further showcase the effectiveness and generality of our approach, we study its use in the context of architecture search for monocular depth estimation. To the best of our knowledge, this is the first attempt at performing NAS for this task. We, therefore, construct a dedicated search space and show that a landmark-regularized NAS algorithm can find novel architectures that improve upon the state of the art in this field.

\section{Related work}
Different from manual designing convolutional neural networks, which have been shown successful in many computer vision tasks~\cite{he2016resnet,guo2020expand,wang2019runet,yu2018smsop},
neural architecture search~(NAS) methods automate the design process and can be categorized into conventional approaches, that obtain architecture performance via stand-alone training~\cite{Zoph2017,Zoph2018,Tan2018,wang2019alphax,real2018regularized,Real2017,wang2020neural}, and weight sharing NAS, where the performance is obtained from one or a few super-nets that encompass all architectures within the search space~\cite{Pham2018,Luo2018,cai2018proxyless,zhao2020fewshot,peng2020creme,you2020greedynas}. 
Motivated by the success of early NAS works, the literature has now branched into several research directions, such as using multi-objective optimization to discover architectures under resource constraints for mobile devices~\cite{tan2019efficient,Tan2018,wu_fbnet:_2018,cai2018proxyless,guo_single_2019,Bender2020tunas}, applying NAS to other computer vision tasks than image recognition~\cite{liu2019autodeeplab,chen2019detnas,li2020adapting,ryoo2020assemblenet}, and using knowledge distillation to eliminate the performance gap between super-net and stand-alone training for 
linear search spaces based on MobileNet~\cite{Cai2020Once,yu2020bignas}. 

In contrast to the diversity of these research directions, super-net training in weight-sharing NAS has remained virtually unchanged since its first appearance in~\cite{li2019random,guo_single_2019,Bender2018}. At its core, it consists of sampling one or few architectures at each training step, and updating the parameters encompassed by these architectures with a small batch of data. 
This approach has been challenged in many ways~\cite{benyahia19overcoming,li2019random,yu2020evalnas,Yang2020NAS}, particular thanks to the introduction of the NASBench series~\cite{radosavovic_network_2019,ying2019bench,dong2020bench102,siems2020bench,Zela2020NAS-Bench-1Shot1,dong2021natsbench} of NAS benchmarks, which provide stand-alone performance of a substantial number or architectures and thus facilitate the analysis of the behavior of NAS algorithms. A critical issue that has been identified recently is the inability of most modern NAS algorithms to surpass simple random search under a fair comparison. In~\cite{yu2020evalnas}, this was traced back to to the low ranking correlation between stand-alone performance and the corresponding super-net estimates. While recent works~\cite{guo_single_2019,chu_fairnas:_2019} have shown that the ranking correlation is high on a MobileNet-based search space, where one only searches for the convolutional operations and the number of channels, others~\cite{yu2020train,Zela2020NAS-Bench-1Shot1} have revealed that the correlation remains low on cell-based NASNet-like search spaces~\cite{Liu2018darts,Luo2018,Zoph2017}, even when carefully tuning the design of the super-net. 

In this paper, we introduce a simple, differentiable regularization term to improve the ranking correlation in weight-sharing NAS algorithms. We show that the regularization term can be used in a variety of weight-sharing NAS algorithms, and that it leads to a consistent improvement in terms of ranking correlation and final search performance.

Our regularization leverages the stand-alone performance of a few architectures. While some contemporary works also use ground-truth architecture performance, our approach differs fundamentally from theirs. Specifically, these methods aim to train a performance predictor, based on an auto-encoder in~\cite{luo2020seminas} or on a graphical neural network in~\cite{tang2020seminas}, and are thus only applicable to weight-sharing NAS strategies that exploit such a performance predictor. By contrast, we add a regularizer to the super-net training loss, which allows our method to be applied to most weight-sharing NAS search strategies. Furthermore, our approach requires an order of magnitude fewer architectures with associated stand-alone performance; in our experiments, we use 30 instead of 300 in~\cite{luo2020seminas} and 1000 in~\cite{tang2020seminas}.

\section{Preliminaries}
\begin{figure*}
    \centering
    \resizebox{\textwidth}{!}{
    \includegraphics{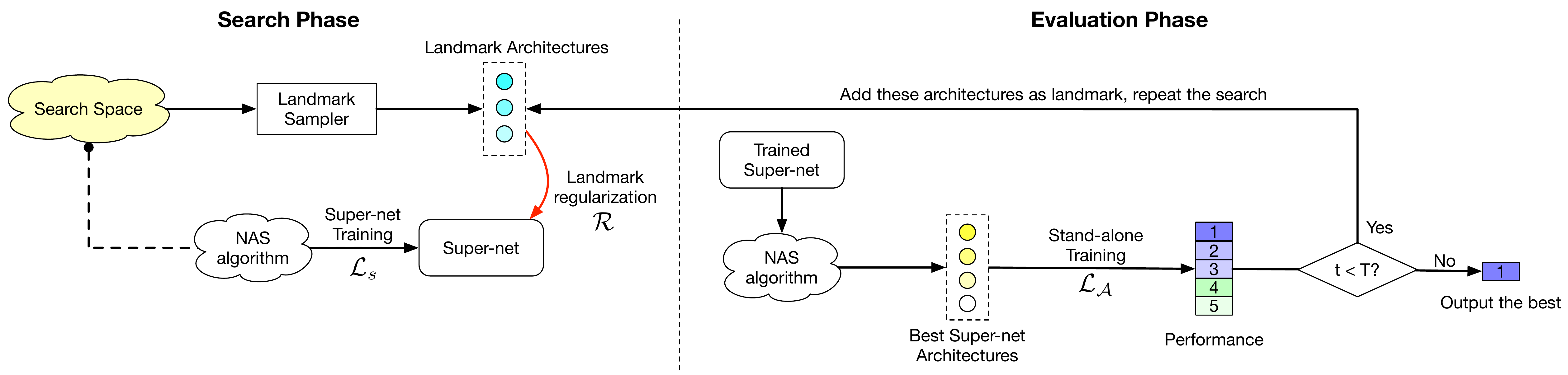}
    }
    \vspace{-0.7cm}
    \caption{
    Overview of our approach. Left: During the search phase, we first sample a set of landmark architectures and obtain their stand-alone performance. We train the super-net with our regularization term such that the landmark ranking is preserved. Right: After a round of training, we sample the best architectures given the current super-net performance and evaluate their stand-alone performance. We add these architectures to the set of landmarks and repeat the
    process for a few iterations.
    }
    \label{fig:method-sketch}
\end{figure*}
\label{sec:method}

We first revisit the basics of super-net training and highlight the ranking disorder problem.

Let $\Omega$ be a search space, defined as a set of $N$ neural network architectures $a_i, i \in [1, N]$.  stand-alone training optimizes the parameters $\theta_{a_j}$ of architecture $a_j$ independently from the other architectures by minimizing a loss function  $\mathcal{L}(x, \theta_{a_j})$, thus yielding the optimal parameters $\theta_{a_j}^*$ for the given training data $x_{train}$. Without weight sharing, NAS then aims to train a search algorithm $S$ to sample architectures $S(\Omega) = \{ a_k \}$ whose stand-alone performance outperforms that of other architectures, that is, $\mathcal{L}(x, \theta_{a_k}^*) < \mathcal{L}(x, \theta_{a_j}^*),\; \forall a_j \notin S(\Omega)$. In its simplest form, i.e., random search, there is no search algorithm to train per se, and one just samples a set of architectures, trains them in a stand-alone fashion, and ranks them to choose the ones with lowest loss. 

With a proper search algorithm, however, NAS without weight sharing requires training and evaluating an impractically large amount of stand-alone architectures \cite{Zoph2017,Zoph2018,wang2020neural,Tan2018,tan2019efficient}
. To circumvent this, weight-sharing NAS strategies construct a super-net $\theta^s$ that encompasses all architectures in the search space. The relative performance of individual architectures sampled from the super-net then acts as an estimate of their relative stand-alone performance. Training is typically formulated as minimizing a joint loss over all architectures that are represented by the super-net:
\begin{align}
\mathcal{L}_s(\theta^s) = \sum_{i=1}^N \mathcal{L}(x, \theta_{a_i}^s),
\end{align}
where the optimization is made tractable by randomly sampling terms from this sum for each update.

In contrast to stand-alone training, parameters overlap between different 
architectures and in general we have that $\theta^s_{a_i} \cap \theta^s_{a_j} \neq \varnothing$. 
Since the parameters shared by $a_i$ and $a_j$ would typically not have the same optimal values in the stand-alone training~\cite{benyahia19overcoming}, the optimal solution of super-net training is \emph{not} the same as that of stand-alone training, and neither is the ranking of the architectures. 

\section{Landmark regularization}
\label{sec:landmark-reg}
We address the issue of low ranking correlation by introducing a simple yet effective approach to regularize super-net training with prior knowledge about the relative performance of individual architectures. To this end, we sample $M << N$ architectures to form a set of landmark architectures, ${\Omega_{L} = \{{a_i}\}_{i=1}^M}$, and obtain their stand-alone performance $\mathcal{L}_{valid}(x_{valid}, \theta_{a_i}^*)$ on validation data, where ${\theta_{a_i}^* = \arg \min \mathcal{L} (x_{train}, \theta_{a_i})}$. We shorten this to $\mathcal{L_{A}}(x, \theta_{a_i}^*)$ to simplify the notation. 
To ensure that the trained super-net is predictive of the 
performance of the stand-alone architectures, we aim to preserve the relative performance of these landmark architectures in the super-net. Formally, we want to enforce that
\begin{align}
\mathcal{L_A}(x, \theta_{a_i}^*) \leq \mathcal{L_A}(x, \theta_{a_j}^*) \Rightarrow \mathcal{L}(x, \theta_{a_i}^s) \leq \mathcal{L}(x, \theta_{a_j}^s),
\label{eq:ordering_loss}
\end{align}
for all pairs of architectures $i, j \in [1, M]$.

To achieve this, we propose a differentiable regularization term that can readily be integrated into standard super-net training procedures. We first sort the landmark architectures in ascending order based on their ground-truth loss:
\begin{align}
\forall i, j \in [1, M],\quad i < j \Leftrightarrow  \mathcal{L_A}(x, \theta_{a_i}^*) \leq \mathcal{L_A}(x, \theta_{a_j}^*),
\label{eq:ordering}
\end{align}
and use this ordering to define a regularization term
\begin{align}
    \mathcal{R}(\theta^s) = \sum_{i = 1}^M  \sum_{j = i+1}^M \max(0,\;\mathcal{L}(x, \theta_{a_i}^s) - \mathcal{L}(x, \theta_{a_j}^s))\;,
    \label{eq:full_reg_loss}
\end{align}
which penalizes deviations from the ordering induced by~\eqref{eq:ordering_loss}.
Since all operations involved in the proposed regularizer are differentiable, it is straight-forward
to implement in existing deep learning frameworks.
Note that the several alternative formulations of this loss are possible. We discuss them in the supplementary
material. 

The landmark-regularized training loss is then given by
\begin{align}
    \mathcal{L}(\theta^s)  = \mathcal{L}_{s}(\theta^s) + \lambda \mathcal{R}(\theta^s)\;,
\end{align}
where $\lambda >0$ is a hyper-parameter. To avoid leakage of validation data into the super-net training, 
we follow \cite{Liu2018darts,Luo2018,cai2018proxyless} and split the training set into two parts. We use one part to evaluate the super-net loss $\mathcal{L}_s$ and the other part to evaluate the regularization $\mathcal{R}$ during training.

\mypara{Computational cost.}
The complexity of evaluating the regularizer discussed above is $O(M^2)$. This factor can have a significant impact on the training time, as the regularizer has to be
evaluated at every training iteration. To reduce this computational burden while still encouraging the super-net to encode a correct architecture ranking, we propose to randomly sample $m$ pairs of landmark architectures $i, j$ at each iteration and evaluate their ranking:
\begin{align}
    \mathcal{R}(\theta^s) = \sum_{i,j}^m\max(0, \mathcal{L}(x, \theta_{a_i}^s) - \mathcal{L}(x, \theta_{a_j}^s))\;.
\end{align}
This reduces the time complexity from $O(M^2)$ to $O(m)$.  We will show empirically that even evaluation with a single pair introduces virtually no degradation of the resulting architecture ranking.

\mypara{Landmark selection.} The choice of the landmark architectures has an impact on the effectiveness of our regularizer. In particular, we would like to use landmarks that cover the complete search space. To promote this, we introduce the diverse landmark sampling strategy described by Algorithm~\ref{alg:landmarks}. We start by randomly sampling a root architecture from the search space. We then generate $M-1$ diverse architectures by mutating the root architecture such that the Hamming distance is larger than a threshold $\tau$. For example, in the DARTS search space~\cite{Liu2018darts}, one architecture is encoded as a sequence, 
where each element represents selecting either a previous node or an operation. Mutating an architecture is then achieved by randomly altering one element, and the Hamming distance between two architectures is computed as the number of unequal elements.

\mypara{Regularization schedule.} 
For all practical applications, the number of landmark architectures will be several orders of magnitude smaller than the total number of architectures 
in the search space. The regularization term thus needs to have high weight to be effective and have a noticeable effect on the training. However, too much 
regularization can negatively impact the training dynamics, especially in the early stages. To alleviate this issue, we propose to enable regularization after a warm-up phase and to gradually increase its influence using a cosine schedule. Specifically, we set the regularization weight at epoch $t$ to
\begin{equation}
\lambda_t = \mathbf{1}_{t > t_w} \cdot \frac{1}{2} \left(1 + \cos{\frac{\pi (t - t_{w})}{t_{total}}}\right)\lambda_{\max},
\end{equation}
where $t_{w}$ denotes the number of warm-up epochs, $t_{total}$ denotes the total number of epochs, and $\lambda_{max}$ denotes the final value for the regularization parameter.

\SetInd{0.5em}{0.5em}
\begin{algorithm}[t]
\small
\SetAlgoLined
\SetAlgoSkip{}
\SetKwInOut{Output}{Output}\SetKwInOut{Input}{Input}
\SetKwFunction{Sample}{RandomSample}
\SetKwFunction{Mutate}{Mutate}

\Input{Search space $\Omega$,  NAS algorithm $\mathcal{S}$, super-net and stand-alone losses  $\mathcal{L}_s$, $\mathcal{L_A}$, distance threshold $\tau$.} 
\BlankLine
\emph{initialize super-net parameters} $\theta^s$

\emph{initialize an empty landmark set} $\Omega_L$

\While{step  $t < T$ }{
\textit{Obtain landmark architectures}

$a_0 \leftarrow \Sample(\Omega, 1)$ 

\While{$|\Omega_L|< M \times T$}{
    $a_t \leftarrow \Mutate(a_0)$
    
    \If{$d_{Hamming}(a_t, a_0) > \tau$}{
        add $a_t$ to $\Omega_L$
    }
}

\ForEach{training step}{
\textit{Train super-net} $\mathcal{L}$ while sampling $m$ pairs

$\{(a_i,a_j)\} \leftarrow $ \Sample$(\Omega_L \times \Omega_L, m)$

$\mathcal{L} = \mathcal{L}_s(\theta^s) + \lambda\mathcal{R}(\theta^s)$

\textit{Train sampler $\mathcal{S}$ if necessary}

}
\textit{Sample architectures to get stand-alone performance}

$ \forall a_j \in \Omega_{t} \leftarrow \mathcal{S}(\Omega)$, obtain $\mathcal{L_A}(x, \theta_{a_j})$\

$\Omega_L \leftarrow \Omega_L \cup \Omega_t$ 
}
\Output{Model $a_t \leftarrow \arg \min_{a \in \Omega_t} \mathcal{L}_{GT}(x, \theta_{a})$}
\caption{Landmark-regularized training.}
\label{alg:landmarks}
\end{algorithm}
\vspace{-0.2cm}

\mypara{Application to existing NAS methods.}
The proposed regularization term is independent of the search algorithm, and thus widely applicable to many different weight-sharing NAS algorithms. We discuss its use in three different classes of NAS strategies, specific instances of which will act as baselines in our experiments.

We categorize weight sharing NAS algorithms into three broad categories according to their interaction with the super-net: i) unbiased architecture sampling algorithms~\cite{li2019random,yu2020evalnas,Bender2018,guo_single_2019,chu_fairnas:_2019} that sample one or a few paths uniformly at random, ii) learning based sampling that favors the most promising architectures given the performance of the current, partially trained super-net~\cite{Pham2018,li2019improving,Luo2018,wang2019alphax,zhao2020fewshot}, and iii) differentiable architecture search that parametrizes the architecture sampling probability as part of the super-net~\cite{Liu2018darts,cai2018proxyless,xie2018snas,nayman2019xnas,Xu2020PC-DARTS:}.

For the first two categories, our method can be directly incorporated into the super-net training to improve its quality, and hence to improve the final search results. For algorithms in the last category, 
the algorithm is usually composed of two distinct phases that are executed alternatingly. In the the first phase the parameters that define the architecture are fixed and only the weights are updated, whereas in the second phase the weights are fixed and the architecture parameters are updated. 
Our regularization term can directly be integrated into the first phase when using discrete architectures in the forward pass as in works that employ the Gumbel-Softmax~\cite{dong2019searching,wu_fbnet:_2018} or binary gates~\cite{cai2018proxyless,xie2018snas}. When using continuous architecture specifications in the forward pass ~\cite{Liu2018darts,Xu2020PC-DARTS:}, we do not have a discrete sub-path to sample and evaluate the regularization term, so additional care has to be taken to incorporate it. We discuss various ways to do this in supplementary material.

\mypara{Multi-iteration pipeline.}
Figure~\ref{fig:method-sketch} depicts the complete training pipeline.  We first sample landmark architectures using our landmark selection strategy and obtain their stand-alone performance. We then train the NAS algorithm with landmark regularization. After a round of training, 
we sample the top $M$ architectures using the trained NAS algorithm, obtain their stand-alone performances, and add these architectures to the set of landmarks.
We proceed training of the super-net with the expanded set of 
landmarks and iterate this process.

In our experiments on different tasks and algorithms, we observed a stable improvement after sampling 3 sets of $M=10$ architectures for a total of $30$ landmarks, which is computationally feasible. Additionally, our algorithm can leverage previously trained models to improve the search by simply adding them to the landmark set.
Considering that search spaces usually encompass billions of architectures, the number of landmarks is negligible.

\section{Experiments}
To validate the landmark regularization we incorporate it into three popular weight-sharing algorithms and evaluate 
them on the task of image classification using the CIFAR-10~\cite{Krizhevsky09cifar} and ImageNet~\cite{imagenet} datasets.
We then discuss a new search space for monocular depth estimation architectures and show that our approach also applies to this new task. Finally, we ablate the key components and hyperparameter choices of our landmark regularization. 

We used PyTorch for all our experiments and follow the evaluation framework defined in~\cite{yu2020evalnas} to ensure a fair comparison with the baseline methods. 
Following~\cite{radosavovic_network_2019,radosavovic2020design}, we shorten the training time from 600 to 100 epochs on CIFAR-10 and from 250 to 50 on ImageNet, which still yields a good prediction quality. 
We release our code at \url{https://github.com/kcyu2014/nas-landmarkreg}.

\mypara{Baselines.} We select single-path one-shot (SPOS) as a representative unbiased architecture sampling algorithm. We use SPOS to train the super-net, followed by an  evolutionary search to select the best models based on the super-net performance~\cite{guo_single_2019}. Among the learning-based architecture sampling methods, we select neural architecture optimization~(NAO)~\cite{Luo2018}, which trains an explicit auto-encoder-based performance predictor. Finally, for differentiable architecture search, we select the gradient-based search using a differentiable architecture sampler~(GDAS)~\cite{dong2019searching}, which has been widely used in other works~\cite{xie2018snas,wu_fbnet:_2018,cai2018proxyless,li2019improving}. See the supplementary material for more details and hyper-parameter settings of the baseline algorithms. %

\mypara{Hyperparameters.} We sample  $M=10$ landmark architectures at each iteration, 
and perform $T=3$ iterations. We sample $m=1$ pairs for each training step and set $\lambda_{max}=10$ in all of our experiments unless otherwise specified. 
The Hamming distance threshold $\tau$ is set according to the configuration of each search space. We train the baselines for the same total number of epochs, to ensure that any performance improvement cannot be attributed to our approach sampling more architectures.

\mypara{Metrics.} We follow~\cite{guo_single_2019,yu2020train} and report the ranking correlation in terms of the \fullskdt~(\skdt). 
We sample 200 architectures randomly to compute this metric for the CIFAR-10 experiments, 90 for the ImageNet experiments, and 20 for the monocular depth estimation experiments. Note that we exclude the landmark architectures from this set to avoid reporting overly optimistic numbers for our approach. Furthermore, following~\cite{yu2020evalnas,dong2020bench102}, we report the mean and best stand-alone performance of the best architectures found over three independent runs.

\subsection{Image classification on CIFAR-10}
\label{subsec:exp-cifar-10}

Since the inception of NAS, CIFAR-10 has acted as one of the main datasets to benchmark NAS performance~\cite{Zoph2018,Pham2018,Luo2018,guo_single_2019,Xu2020PC-DARTS:,you2020greedynas}. We utilize two search spaces, NASBench-101 and NASBench-201, for which the stand-alone performance of many architectures is known. %

\begin{table}[t]
    \centering
    \resizebox{\linewidth}{!}{
    \begin{tabular}{l|ccccc}
    \toprule
    & \multicolumn{5}{c}{NASBench-101} \\
    \cmidrule{2-6}
    Model     &  S-KdT  &  Mean Acc. &  Best Rank & Best Acc. & Cost \\
    \midrule
    SPOS & 0.267 $\pm$ 0.02 &  91.02 $\pm$ 0.52 & 38953 & 92.82 & 11.89 \\
    SPOS+Ours & 0.347 $\pm$ 0.03  & 92.48 $\pm$ 0.51 & 27697  & 92.99 & 14.92  \\
    \midrule
    NAO & 0.329 $\pm$ 0.11 & 90.56 $\pm$ 0.88 &  131969 &  91.60 & 15.46 \\
    NAO+Ours & 0.457 $\pm$ 0.03 &  92.23 $\pm$ 1.32 & 9313 & 93.34 & 21.04 \\
    \midrule
    \midrule
    & \multicolumn{5}{c}{NASBench-201} \\
    \cmidrule{2-6} 
    Model &  S-KdT  &  Mean Acc. & Best Rank  & Best Acc. & Cost \\
    \midrule
    SPOS & 0.771 $\pm$ 0.04 & 87.66 $\pm$ 4.95 & 3383 & 92.30 & 6.26 \\
    SPOS+Ours & 0.802 $\pm$ 0.02 & 92.08 $\pm$ 0.37 & 2557 & 92.53 & 7.23 \\
    \midrule
    GDAS &  0.691 $\pm$ 0.01 & 93.58 $\pm$ 0.12 & 463 & 93.48 & 14.42 \\
    GDAS+Ours & 0.755 $\pm$ 0.01 & 93.98 $\pm$ 0.09 &  109 & 93.84 & 16.28 \\
    \midrule
    NAO  & 0.653 $\pm$ 0.05  &  91.75 $\pm$ 1.52  & 649 & 93.35 & 3.51 \\
    NAO+Ours & 0.758 $\pm$ 0.05 &  92.84 $\pm$ 0.71 & 179 & 93.75 & 4.20 \\ 
    \bottomrule
    \end{tabular}
    }
    \vspace{-0.1cm}
    \caption{Results on NASBench-101 and NASBench-201. We report the S-KdT at the end of training, the mean stand-alone accuracy of the searched architectures, the best rank, and the best accuracy. 
    Each method was run 3 times. We also report the search cost in hours on Tesla-V100~(32Gb).
    }
    \label{tab:cifar10-nasbench}
\end{table}
\mypara{NASBench-101.} NASBench-101~\cite{ying2019bench} is a cell search space that contains 423,624 architectures with known stand-alone accuracy on CIFAR-10. It is the largest tabular benchmark search space to date. We use the implementation of~\cite{yu2020evalnas} to benchmark the performance of SPOS and NAO. 
We do not report the results of GDAS
on this search space, as the matrix-based configuration used in NASBench-101 is not amenable to differentiable approaches~\cite{ying2019bench,Zela2020NAS-Bench-1Shot1}. 
We set the Hamming distance threshold to $\tau=5$.

As shown in Table~\ref{tab:cifar10-nasbench} (top), landmark regularization improves the ranking correlation (\skdt) from 0.267 to 0.347 for SPOS, and from 0.329 to 0.457 for NAO. This translates to a 1-2\% improvement in terms of mean stand-alone accuracy over three runs. The best architecture discovered on this search space, thanks to our regularizer, ranks 9313-th which corresponds to the top 2\% of architectures. Note that NAO without landmark regularization consistently got trapped in local minima, leading to its best architecture being only in the top 30\%. 

\mypara{NASBench-201.} 
\begin{figure}
    \centering
    \resizebox{\linewidth}{!}{
    \includegraphics{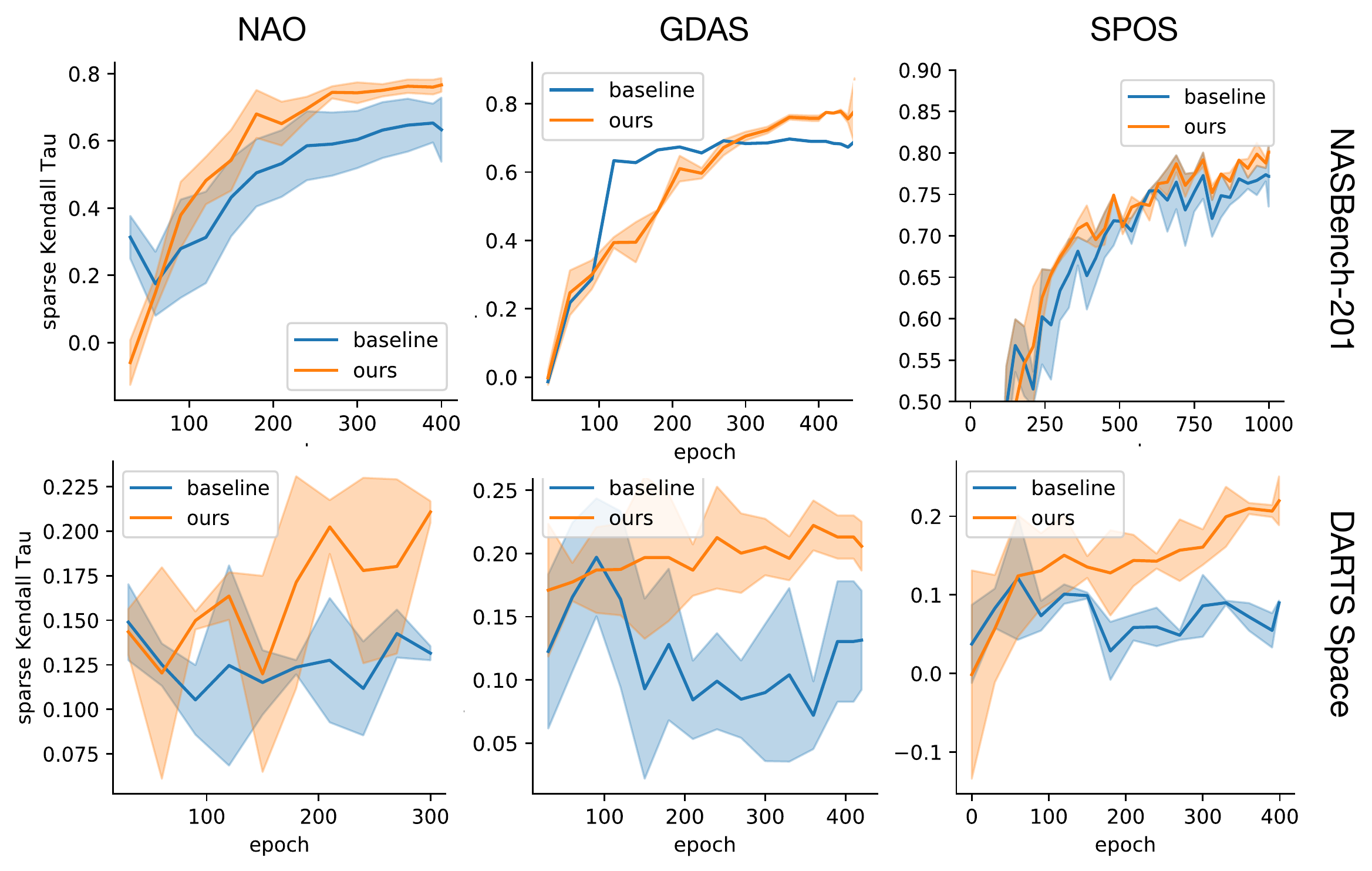}
    }
    \caption{Evolution of the S-KdT of three NAS algorithms on two search spaces. Landmark regularization significantly improves the ranking correlation of the super-net in all cases. 
    }
    \label{fig:skdt-cifar10}
\end{figure}
We report results on the NASBench-201 cell search space~\cite{dong2020bench102}, where each cell is a fully connected graph with 4 nodes. Each edge contains 5 searchable operations.
This yields a total of 15,625 architectures. We set $\tau=5$ and benchmark all three baseline algorithms on this space. As shown in Table~\ref{tab:cifar10-nasbench} (bottom), landmark regularization consistently improves the ranking correlation (\skdt) across all three methods. We also observe an improvement in mean accuracy  of more than 4\% with SPOS.
The best architecture is obtained by GDAS with landmark regularization and ranks 109-th. This corresponds to the top 0.7\% architectures across the search space. In Figure~\ref{fig:skdt-cifar10} (top), we plot the mean \skdt~as super-net training progresses. We can see that the regularization improves the ranking correlation by a significant margin, especially towards the end of training.
Note that we extend our experiments to CIFAR-100 and ImageNet16-120 of NASBench-201 in supplementary material.

\begin{table}
    \centering
    \resizebox{\linewidth}{!}{
    \begin{tabular}{l|ccccc}
    \toprule
    Model     &  S-KdT  & Mean Acc. &  Params & Best Acc. & Cost \\
    \midrule
    SPOS &  0.058 $\pm$ 0.010 &  92.80 $\pm$ 0.03 & 5.082M & 92.88 & 12.53 \\
    SPOS+Ours & 0.206 $\pm$ 0.018 & 93.41 $\pm$ 0.43 & 2.181M & 93.84 & 15.96 \\
    \midrule
    GDAS & 0.176 $\pm$ 0.014   & 90.48 $\pm$ 2.95  &3.418M & 93.43 & 9.27 \\
    GDAS+Ours & 0.209 $\pm$ 0.001  & 94.32 $\pm$ 0.28 & 2.540M & 94.60 & 13.24 \\
    \midrule
    NAO & 0.102 $\pm$ 0.018  &  92.93 $\pm$ 0.87 & 5.080M & 93.03 & 19.72 \\
    NAO+Ours & 0.231 $\pm$ 0.012 & 93.53 $\pm$ 0.43 & 2.184M & 93.78 & 28.18 \\
    \bottomrule
    \end{tabular}}
    \caption{Results on the DARTS search space on {CIFAR-10}. Our best model (GDAS+Ours) surpasses the state-of-the-art model of~\cite{Xu2020PC-DARTS:} (94.02\% accuracy with 3.62M parameters) with 30\% fewer parameters. %
    }
    \label{tab:darts-cifar10}
\end{table}
\mypara{DARTS search space.} The NASBench search spaces are relatively small. To evaluate our approach in a more realistic scenario, we make use of the DARTS search space~\cite{Liu2018darts,xie2018snas,Xu2020PC-DARTS:,nayman2019xnas} which spans $3.3\times 10^{13}$ architectures and is commonly used to evaluate real-world NAS performance. In contrast to NASBench, for which we could query the existing stand-alone performances, here, we need to train the discovered architecture from scratch. 
To compute the ranking correlation, we rely on the 5,000 pre-trained models of~\cite{radosavovic_network_2019} from which we randomly sample 200 as before. 

We report our results in Table~\ref{tab:darts-cifar10}. We observe a clear improvement in terms of both \skdt~and mean accuracy over three independent searches across all three algorithms. Interestingly, the best model obtained using our ranking loss can surpass the baseline models by almost 1\% with only around 50\% of the parameters. The best model from GDAS with our regularizer surpasses the state-of-the-art model on this space~\cite{Xu2020PC-DARTS:} by 0.58\% with 30\% fewer parameters. %

\subsection{Image classification on ImageNet}
\begin{figure*}[t]
    \centering
    \resizebox{\textwidth}{!}{\includegraphics{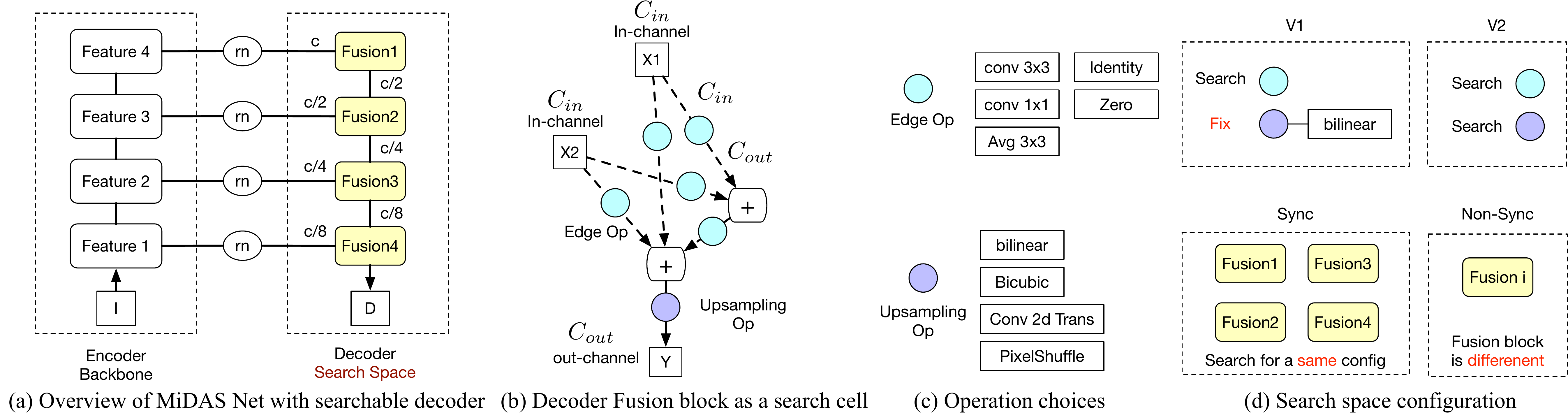}}
    \vspace{-0.5cm}
    \caption{Monocular depth estimation search space. (a) We modify MiDaS~\cite{Ranftl2020} to construct the search space. We keep the backbone unchanged and search for fusion blocks in the decoder branches. (b) To define a fusion block, we model the input from the backbone and from the preceding fusion block as two nodes. We add two feature nodes, which sum up all previous inputs. The nodes are connected by edges, which represent the searchable operations.
    The final output node $Y$ takes the output of the last feature node and applies a potentially searchable upsampling operation. (c) Each edge, except the output edge, represents an \textit{Edge Op (blue)} that contains five operations to choose from, while an upsampling edge \textit{(purple)} contains four. (d) We propose four sub-space configurations: `V1' indicates that we only search edge operations and fix the upsampling operator to bilinear upsampling. `V2' includes a search over the upsampling operators. `Sync' indicates that all fusion blocks share the same configuration, while `Non-sync' allows them to differ.
    }
    \label{fig:monodepth-space}
\end{figure*}

\begin{table}[]
    \centering
    \resizebox{\linewidth}{!}{
    \begin{tabular}{l|cc|ccc}
    \toprule
    Model     &  S-KdT  &  Mean Top-1 & Params & Best Top-1 (50/250) & Cost \\
    \midrule
    SPOS &0.210 $\pm$ 0.010 &  64.57 $\pm$ 3.30 & 4.579M & 67.88 / 73.69 & 7.29 \\
    SPOS+Ours & 0.267 $\pm$ 0.018   &   67.38 $\pm$ 0.92 & 4.766M & 68.61 / 74.58 & 9.39 \\
    \midrule
    GDAS & 0.247 $\pm$ 0.012 &  67.50 $\pm$ 0.26 & 5.076M &67.26  / 74.03 & 9.13 \\
    GDAS+Ours & 0.272 $\pm$ 0.023   & 68.82 $\pm$ 0.24 & 5.073M & 68.36 / 74.82 & 10.39 \\
    \midrule
    NAO & 0.253 $\pm$ 0.006  & 67.70 $\pm$ 0.43 &  4.675M &68.21 / 73.71 & 8.39 \\
    NAO+Ours &  0.279 $\pm$ 0.003 & 68.89 $\pm$ 0.58 & 4.488M &69.58 /  74.92 & 11.21\\
    \bottomrule
    \end{tabular}}
    \caption{Results on ImageNet. We report mean top-1 accuracy over 3 runs after 50 epochs and best top-1 accuracy 
    after 50 and 250 epochs, respectively.}
    \label{tab:imagenet-nds}
\end{table}

To further evidence the effectiveness of our method, we move to ImageNet classification. For evaluation, we pick the best model of three independent runs
predicted by each NAS algorithm and train them from scratch on the entire ImageNet training set for 50 epochs. We follow the setup of~\cite{Xu2020PC-DARTS:} and use stochastic gradient descent with a linear learning rate scheduler, which linearly increases from 0.1 to 0.5 in the first five epochs, and decreases to 0 over the remaining 45 epochs. We use a weight decay of $3\text{e-}4$ and a label smoothing coefficient of $0.1$ for all models. For this dataset, we use the popular DARTS search space, which provides 120 architectures trained for 50 epochs~\cite{radosavovic_network_2019}. We split these architectures into 90 to report the sparse Kendall-Tau evaluation metric, and 30 for landmark sampling. We train the super-net with only 15\% of the training dataset as in~\cite{Xu2020PC-DARTS:}. As the test data is not public, we report the top-1 validation accuracy as a metric for stand-alone training.

\mypara{Results.} Table~\ref{tab:imagenet-nds} shows that landmark regularization consistently improves the three baselines. Overall, the best model is found by NAO with landmark regularization and achieves 74.92\% top-1 accuracy, which outperforms the best baseline model by more than 1\%. This further evidences the effectiveness of our regularization and its robustness across datasets.

\subsection{Monocular depth estimation}

To showcase the generality of our approach, we apply our landmark-regularized NAS to the task of monocular depth
estimation. Monocular depth estimation aims to predict pixel-wise depth from a single RGB image. Different paradigms
have emerged on how to train single-image depth predictors, ranging from fully supervised training \cite{Alhashim2018,Xian_2018_CVPR,Ranftl2020}, to self-supervised approaches \cite{Godard2019}.  We follow the supervised paradigm and use the loss function proposed by Ranftl \etal \cite{Ranftl2020} to search for an architecture on the ReDWeb \cite{Xian_2018_CVPR} dataset.

\mypara{Search space.}
Figure~\ref{fig:monodepth-space} gives a detailed overview of the structure of our search space.
Figure~\ref{fig:monodepth-space}~(a) shows the structure of  a traditional monocular depth estimation network. It is composed of a backbone network that acts as a feature extractor, typically pre-trained on ImageNet, followed by decoder fusion blocks that aggregate multi-scale information into a final prediction. Since using a pre-trained high-capacity network has been shown to be of high importance for final performance~\cite{Alhashim2018,Ranftl2020}, its architecture
is fixed. We thus propose to search for the fusion blocks that define the decoder. 

As depicted in Figure~\ref{fig:monodepth-space}~(b), each fusion block is a search cell and takes the output of its preceding fusion block and the features from the backbone network as input. As the first fusion block does not have a predecessor, we simply duplicate the features from the encoder. 
Figure~\ref{fig:monodepth-space}~(c) shows the possible searchable operations, whereas (d) illustrates four possible configurations of our search space. 
We report the total number of architectures $|\Omega|$ for each search space in Table~\ref{tab:mono-depth}. 
While the search space is relatively simple, it is large enough to exhibit the problem of ranking disorder.

\begin{table}[!t]
  \centering
  \resizebox{\linewidth}{!}{
  \begin{tabular}{c|c|rr|rr}
    \toprule
    \multirow{2}{*}{Search space} & \multirow{2}{*}{Method} & \multicolumn{2}{c|}{Sync} & \multicolumn{2}{c}{Non-sync} \\
    \multicolumn{1}{l|}{}&\multicolumn{1}{c|}{} & \multicolumn{1}{c}{s-KdT} & \multicolumn{1}{c|}{Best Val. loss} & \multicolumn{1}{c}{S-KdT} & \multicolumn{1}{c}{Best Val. loss} \\
    \midrule
    \multirow{3}{*}{V1}
     & &\multicolumn{2}{c|}{$|\Omega| = 3,125$} & \multicolumn{2}{c}{$|\Omega| = 9.5 \times 10^{13}$} \\
     & SPOS &      0.751 $\pm$ 0.003 & 0.0960 $\pm$ 0.001 &  0.732 $\pm$ 0.008  &  0.0973 $\pm$ 0.002 \\
     & SPOS+\textit{Ours}  & 0.781 $\pm$ 0.002 & 0.0958 $\pm$ 0.001 &  0.867 $\pm$ 0.044 & 0.0974 $\pm$ 0.001 \\
     \midrule
    \multirow{3}{*}{V2}
    & &\multicolumn{2}{c|}{$|\Omega| = 12,500$} & \multicolumn{2}{c}{$|\Omega| = 2.4 \times 10^{16}$} \\
     & SPOS & 0.401 $\pm$ 0.010  & 0.0957 $\pm$ 0.001 & 0.611 $\pm$ 0.004 & 0.0973 $\pm$ 0.001 \\
     & SPOS+\textit{Ours}  & 0.555 $\pm$ 0.026 & \textbf{0.0936 $\pm$ 0.000} & 0.681 $\pm$ 0.002 & 0.0964 $\pm$ 0.001 \\
     \bottomrule
  \end{tabular}
  }
  \caption{Results on the RedWeb validation set. The performance achieved by \cite{Ranftl2020} is 0.0942 (lower is better).}
  \label{tab:mono-depth}
\end{table}

\mypara{Results.}
We use the single-path one-shot algorithm to benchmark the influence of our regularization term on this task.
We run $T=2$ iterations, and sample 10 architectures in the first iteration. After the first round 
of training, we pick the three top models, obtain their stand-alone performance, and add them to the set of landmarks before we perform training for a second iteration. In addition to \skdt, we report the scale- and shift-invariant loss \cite{Ranftl2020} on the validation set as the performance metric.

Our results in Table~\ref{tab:mono-depth} indicate that our method consistently yields an improvement in terms of \skdt~for all four configurations of the search space. Our best model improves upon the state-of-the-art handcrafted architecture of~\cite{Ranftl2020} in terms of the final performance. %

\subsection{Ablation studies}
We finally provide an analysis of different aspects of our approach. We first evaluate the influence of the hyper-parameter $\lambda$ and different regularization schedules.
We further study the robustness of our method to the landmark sampling distance $\tau$ and the number of iterations $T$. We  conduct the ablations under the experimental setting described in Section~\ref{subsec:exp-cifar-10} and use SPOS as baseline.

\begin{table}[t]
    \centering
    \resizebox{\linewidth}{!}{\includegraphics[]{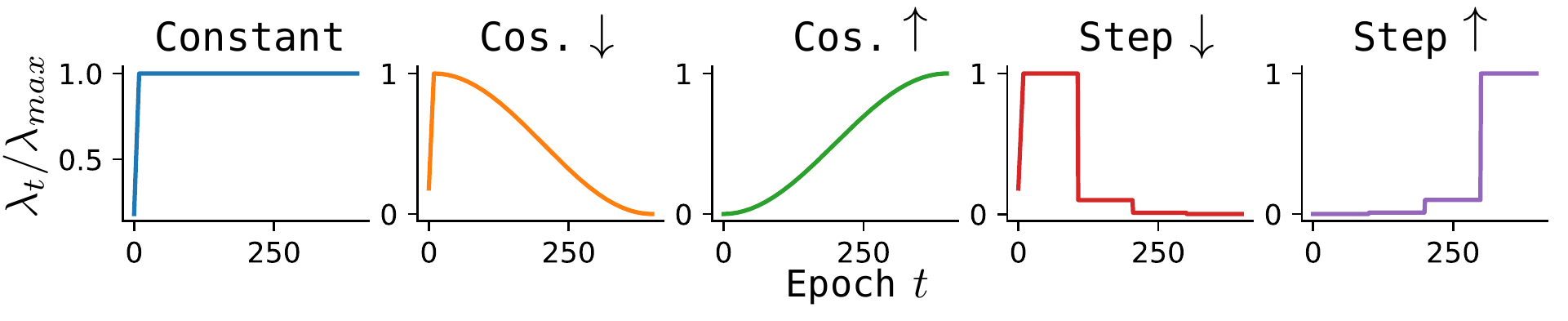}}
    \vspace{0.1cm}
    \resizebox{\linewidth}{!}{
    \begin{tabular}{c|ccccc | ccc}
    \toprule
     \multicolumn{1}{c}{}   & \multicolumn{5}{c|}{Schedule} & \multicolumn{3}{c}{$\lambda_{max}$} \\
        \cmidrule{1-5} \cmidrule{6-9}
         &  Const & Cos $\downarrow$   & Cos $\uparrow$ & Step $\downarrow$   & Step $\uparrow$  & 1 & 10 & 100 \\
        \midrule
        S-KdT & 0.753 & 0.712 & 0.803 & 0.709 & 0.785 & 0.789 & 0.803 & 0.793  \\
        Mean Acc. & 91.13 & 91.38 & 92.08 & 91.02 & 91.31 & 91.89 & 92.08 & 91.97  \\
        \bottomrule
    \end{tabular}
    }
    \vspace{-0.2cm}
    \captionof{table}{Influence of the regularization parameter $\lambda$. Left: Different schedules (\cf top plots) to modify the strength of regularization throughout training. Right: Influence of $\lambda_{max}$ with the increasing cosine schedule.}
    \label{tab:coef-ablation}
    \vspace{0.4cm}
    \resizebox{0.8\linewidth}{!}{
    \begin{tabular}{c|ccc|ccc}
    \toprule
     \multicolumn{1}{c}{}   & \multicolumn{3}{c|}{SPOS} & \multicolumn{3}{c}{SPOS+\textit{Ours}} \\
        \cmidrule{1-3} \cmidrule{4-7}
        Iteration & T=1 & T=3 & T=10 & T=1 & T=3 & T=10  \\
        \midrule
        S-KdT & 0.763 & 0.771 & 0.758 & 0.760 & 0.802 & 0.811  \\
        Mean Acc. & 91.13 & 91.48 & 91.78 & 91.29 & 92.08 & 92.17  \\
        \bottomrule
    \end{tabular}
    }
    \vspace{-0.2cm}
    \captionof{table}{Influence of the number of iterations $T$ on NASBench-201.}
    \label{tab:iteration-ablation}
    \vspace{0.3cm}
    \resizebox{0.9\linewidth}{!}{
    \begin{tabular}{c|ccccccc}
    \toprule
        $m$ Pair(s) & 0 &1 & 2 & 10 & 20 & 50 & 100 \\
        \midrule
        S-KdT & 0.771 & 0.803 & 0.801 & 0.805 & 0.812& 0.807 & 0.791 \\
        Mean Acc. & 87.66 & 92.08 & 92.12 & 92.13 & 92.10 & 92.11 & 92.00 \\
        \bottomrule
    \end{tabular}
    }
    \vspace{-0.2cm}
    \captionof{table}{Influence of the stochastic approximation of Eq.~\ref{eq:full_reg_loss}. We randomly sample $m$ pairs from $30$ landmarks during each training step .}
    \label{tab:pairs-ablation}
\end{table}
\mypara{Loss coefficient $\lambda$ and scheduler.}
In Table~\ref{tab:coef-ablation}, we ablate five different coefficient schedulers: constant regularization throughout training, two schedulers that gradually decrease the regularization, and two schedulers that gradually increase regularization (\cf Table~\ref{tab:coef-ablation} (top)). For the constant and decreasing schedulers, we gradually increase the loss from 0 to $\lambda_{max}$ linearly in the first 10 epochs to avoid an abrupt change of the loss.
As shown in Table~\ref{tab:coef-ablation}, the cosine increasing scheduler, with $\lambda_{max} = 10$, yields the best results. We used this strategy in all our experiments.

\mypara{Iterations.}
We investigate the impact of the number of iterations $T$ in Table~\ref{tab:iteration-ablation}. We first pre-train the super-net for 250 epochs and then continue training for another 150 epochs per iteration $T$ both with and without landmark regularization. Table~\ref{tab:iteration-ablation} indicates that our method improves the results when increasing the number of iterations, while the performance of the baseline does not increase. We selected $T=3$ for our experiments as it strikes a balance between efficiency and accuracy.

\mypara{Sampling in loss computation.} In Table~\ref{tab:pairs-ablation}, we show that sampling a single pair of architectures per iteration to compute the regularization term is sufficient. Using more pairs does not improve the ranking correlation. We hypothesize this to be due to the fact that, overall, super-net training undergoes thousands of steps, thus providing a good coverage of all possible combinations of landmark architectures when the landmark set is small.

\begin{figure}
    \centering
    \resizebox{0.9\linewidth}{!}{\includegraphics{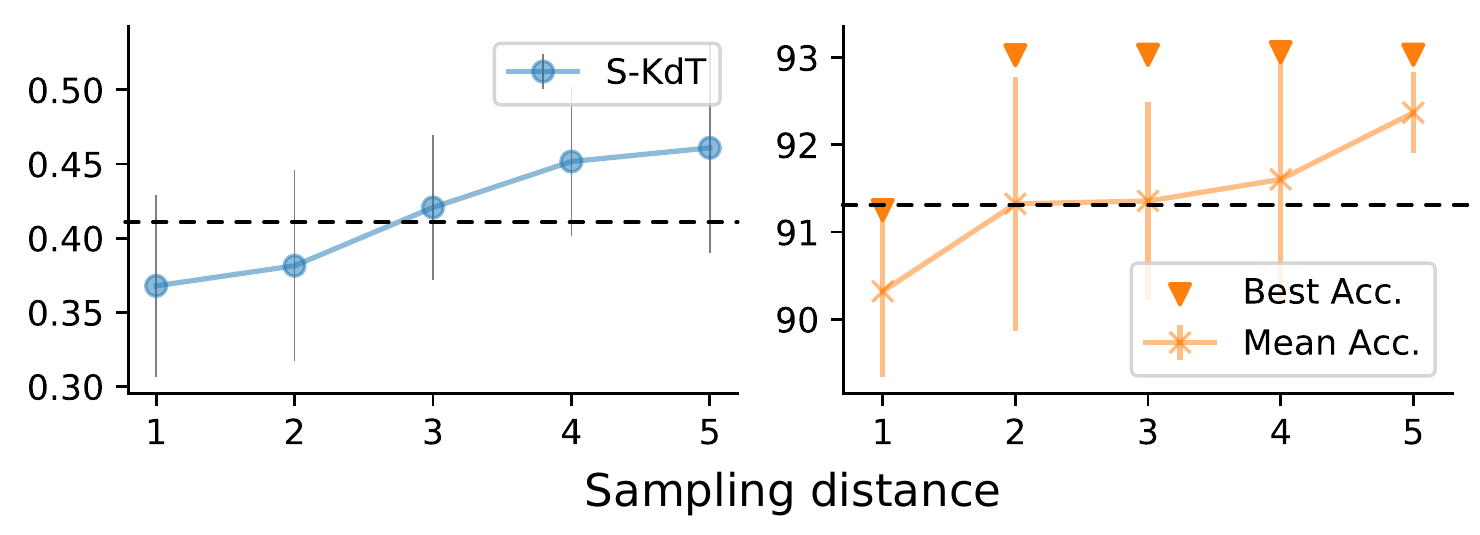}}
    \vspace{-0.2cm}
    \caption{Comparison of different sampling distances $\tau$. The black, dashed line indicates the baseline performance. }
    \label{fig:ablation-distance}
\end{figure}
\mypara{Influence of the sampling distance.}
We finally evaluate the importance of the distance threshold $\tau$ in our landmark sampler. We first pre-train the super-net with SPOS on NASBench-201 for 150 epochs, then train for 50 epochs with landmark regularization, where we sample landmarks with varying distances $\tau$. We repeated this experiment 3 times and report the average \skdt~as well as the mean accuracy of the discovered architectures. Figure~\ref{fig:ablation-distance} shows that the performance degrades if $\tau$ is chosen too small. Performance gradually improves as $\tau$ increases. This highlights the importance of a diverse set of landmarks. Note that our sampling strategy does not require knowledge about the stand-alone performance and thus is applicable to new NAS search spaces.

\section{Conclusion}
We have presented a simple yet effective approach to leverage a few landmark architectures to guide the super-net training of weight-sharing NAS algorithms towards a better correlation with stand-alone performance. Our strategy is applicable to most NAS algorithms and our experiments have shown that it consistently improves both the ranking correlation between the super-net and stand-alone performance as well as the final performance across three different search algorithms and three different tasks. Additionally, our approach can leverage the information from previously trained stand-alone models to improve NAS performance. In the future, we will focus on developing a more advanced landmark sampling strategy.


\end{document}